# Exploring the Efficacy of Large Language Models in Summarizing Mental Health Counseling Sessions: A Benchmark Study

Prottay Kumar Adhikary[1], Aseem Srivastava[2], Shivani Kumar[2], Salam Michael Singh[1], Puneet Manuja[3], Jini K Gopinath[3], Vijay Krishnan[4], Swati Kedia[5], Koushik Sinha Deb[5], Tanmoy Chakraborty[1,6,*]

[1]Department of Electrical Engineering, Indian Institute of Technology Delhi, India
[2]Department of Computer Science & Engineering, Indraprastha Institute of Information Technology Delhi, India
[3]YourDOST, Karnataka, India
[4]Department of Psychiatry, All India Institute of Medical Sciences, Rishikesh, India
[5]Department of Psychiatry, All India Institute of Medical Sciences, New Delhi, India
[6]Yardi School of Artificial Intelligence, Indian Institute of Technology Delhi, India

* Corresponding author (tanchak@iitd.ac.in)

## Abstract

**Background:** Comprehensive summaries of sessions enables an effective continuity in mental health counseling, facilitating informed therapy planning. Yet, manual summarization presents a significant challenge, diverting experts' attention from the core counseling process. Leveraging advancements in automatic summarization addresses this issue, offering mental health professionals accessibility and efficiency by streamlining the summarization of lengthy therapy sessions. However, existing approaches often overlook the nuanced intricacies inherent in counseling interactions.

**Objective:** This study evaluates the effectiveness of state-of-the-art Large Language Models (LLMs) in selectively summarizing various components of therapy sessions through aspect-based summarization, aiming to benchmark their performance.

**Methods:** We introduce MentalCLOUDS, a counseling-component guided summarization dataset. This benchmarking dataset consists of 191 counseling sessions with summaries focused on three distinct counseling components (*aka* counseling aspects). Additionally, we assess the capabilities of 11 state-of-the-art LLMs in addressing the task of component-guided summarization in counseling. The generated summaries are evaluated quantitatively using standard summarization metrics and verified qualitatively by mental health professionals.

**Results:** Our findings demonstrate the superior performance of task-specific LLMs such as MentalLlama, Mistral, and MentalBART in terms of standard quantitative metrics such as Rouge-1, Rouge-2, Rouge-L, and BERTScore across all aspects of counseling components. Further, expert evaluation reveals that Mistral supersedes both

MentalLlama and MentalBART based on six parameters — affective attitude, burden, ethicality, coherence, opportunity costs, and perceived effectiveness. However, these models share the same weakness by demonstrating a potential for improvement in the opportunity costs, and perceived effectiveness metrics.

**Conclusions:** While LLMs fine-tuned specifically in the mental health domain exhibit better performance based on automatic evaluation scores, expert assessments indicate that these models are not yet reliable for clinical applications. Further refinement and validation are necessary before their implementation in practice.

## Introduction

Counseling refers to a relationship between a professional counselor and individuals, families or other groups that empowers the client to achieve mental health, wellness, education and career goals. Specifically, in those with psychological or interpersonal difficulties, mental health counseling may be seen as a key helping intervention. Counseling sessions embrace a client-centered approach, fostering an environment of trust and exploration. These sessions delve deep into personal experiences, where clients share intimate details while therapists navigate the dialogue to cultivate a safe and supportive space for healing. Discussions within these sessions span a wide range of topics, from recent life events to profound introspections, all of which contribute to the therapeutic journey. An important aspect of the counseling process lies in the documentation of counseling notes (summary of the entire session), which is essential for summarizing client stressors and therapy principles. Session notes are pivotal in tracking progress and in guiding future sessions. However, capturing the intricacies of these conversations poses a formidable challenge, demanding training, expertise and experience of mental health professionals. These summaries distill key insights, including symptom and history exploration, patient discovery, and reflections, while filtering out non-essential details. Yet, the need for meticulous record-keeping can sometimes detract from the primary focus of therapy. Maintaining a seamless flow of conversation is paramount in effective therapy, where any disruption can impede progress. To streamline this process and ensure continuity, automation emerges as a promising solution for the counseling summarization task. While advancements in Artificial Intelligence (AI) have revolutionized document summarization, the application of these technologies to mental health counseling remains relatively unexplored.

Previous studies [1–3] recognized the potential of counseling summarization in optimizing therapeutic outcomes. However, existing models often overlook the unique nuances inherent in mental health interactions. Standard counseling dialogues, using reflective listening, involve the identification of current issues, developing a biopsychosocial conceptualization including past traumas, coping strategies etc. and chalk out treatment plans. It also includes discussion on between-session issues as well as crisis, if any. An effective counseling summary should selectively capture information pertinent to each of these categories while eliminating extraneous details.

Despite the demonstrated capabilities of Large Language Models (LLMs) in various domains, research in mental health counseling summarization is scarce. One major obstacle is the lack of specialized datasets tailored to counseling contexts. To bridge this gap, we embark on a two-pronged approach: (i) creating a novel counseling-component guided summarization dataset, called MentalCLOUDS, and (ii) evaluating state-of-the-art LLMs on the task of component-guided counseling summarization. Through these efforts, we aim to propel the integration of AI technologies into mental health practice, ultimately enhancing the quality and accessibility of therapeutic interventions.

**Related Work**
Summarizing counseling conversations eases the continuity of sessions and devises comprehensive therapy plans. However, analyzing these interactions manually is an arduous task. To address this challenge, advancements in Artificial Intelligence and Natural Language Processing (NLP), particularly in summarization techniques, offer a promising solution. Summarization tasks can be approached via an extractive [4] or an abstractive [5] view point. Extractive summarization involves identifying the most relevant sentences from an article and systematically organizing them. Given the simplicity of the approach, the resultant extractive summaries are often less fluent. On the other hand, abstractive summarization extracts important aspects of a text and generates more coherent summaries. By employing summarization, therapists can access concise recap of sessions, sparing them the need to shift through lengthy dialogues. While summarization has been a long-studied problem in NLP [6], recent attention has shifted towards aspect-based summarization, a method that focuses on generating summaries pivoted on specific points of interest within documents.

Chen et al. [1] propose a retrieval-based medical document summarization approach where the user query is fine-tuned using a medical ontology. Their method is limited due to its overall primitive design. Additionally, Konovalov et al. [7] highlight the importance of identifying emotional reactions and "early counseling" components. Strauss et al. [8] use machine learning approaches to automate the analysis of clinical forms and envision using machine learning in mental health to a certain extent. Furthermore, research on Major Depressive Disorder (MDD) [9] underscores the significance of identifying crucial indicators from patient conversations such as age, anxiety levels, and long episode duration in the choice of the appropriate level of antidepressant, guiding subsequent sessions and prescriptions. Subsequently, the prescribed antidepressants undergo monitoring to assess the patient's response. This concept identifies crucial indicators from the patient's conversations with the therapist and guides subsequent follow-up sessions based on the patient's historical interactions and prescriptions. Deep learning approaches, such as the use of Recurrent Neural Networks (RNN) and Long Short-Term Memory (LSTM) , have been employed to predict 13 predefined mental illnesses based on neuropsychiatric notes. On average, these notes contain 300 words about the patient's present illness and events associated with it, followed by a psychiatric review system that mentions the mental illness related to the patient. Chen et al. [11] propose an extractive summarization approach using the BERT model [12] to reduce doctors' efforts in analyzing tedious amounts of diagnosis reports. However, there remains a notable gap in effectively capturing medical information in summaries.

Additionally, some contemporary work utilize authentic mental health records to create synthetic datasets [13]. Afzal et al. [14] report summarization of medical documents to identify PICO (Patient/Problem, Intervention, Comparison, and Outcome). Manas et al. [15] propose an unsupervised abstractive summarisation where the domain knowledge from the patient health care questionnaire (PHQ-9) is used to build knowledge graphs to filter relevant utterances. A two-step summarisation is devised by Zhang et al. [16] wherein partial summaries are initially consolidated, and the final summary is generated by fusing these chunks. Additionally, Zafari and Zulkernine [17] demonstrate an online application built using information extraction and annotation tailored to the medical domain.

For dialogue summarization, abstractive summarization has been the de facto due to its ability to capture critical points coherently. Nallapati et al. [18] employ an encoder-decoder-based abstractive summarization method, which is further improved via the attention mechanism [19]. Subsequently, See et al. [20] introduce a hybrid approach of extractive and abstractive summarization. Chen and Bansal [2] propose a reinforcement learning-based approach as a mixture of extractive and abstractive approaches for summarization, wherein the emphasis is given to the redundancy reduction in the extracted utterances from the conversation. Recent research reveals the dependence of specific utterances in the salient sentence extraction from the conversation utterances. In this regard, Narayan et al. [3] perform a topic distribution based on Latent Dirichlet Allocation [21]. Subsequently, Song et al. [22] segregate the utterances into three labels viz. problem description, diagnosis, and others. In medical counseling, Quiroz et al. [23] and Krishna et al. [24] adopt the method of selecting significant utterances for summarizing medical conversations.

In aspect-based summarization, instead of an overall summary of the entire document, summaries at different aspect levels are made based on specific points of interest. These aspects could be movie reviews [25–28] or summarization guided by different domains [29,30] where the documents or the segments of the documents are tagged with these aspects. Hayashi et al. [31] release a benchmarking dataset on multi-domain aspect-based summarization where they annotate 20 different domains as aspects using the section titles and boundaries of each article from Wikipedia articles. Frermann et al. [29] report an aspect-based summarization of the news domain. Their method can segment the documents by aspect, and the model can generalize from the synthetic data to natural documents. The study further reveals the models' efficacy in summarizing long documents. Recently, aspect-based summarization has garnered considerable traction;

however, the dataset is limited. Subsequently, Yang et al. [32] released a large-scale, high-quality dataset on aspect-based summarization from Wikipedia. The dataset contains approximately 3.7 million instances covering around 1 million aspects sourced from 2 million Wikipedia pages. Apart from the dataset release, the authors also benchmark it on the Longformer-Encoder-Decoder (LED) [33] model where they perform zero-shot, few-shot and fine-tuning on seven downstream domains where data is scarce. Joshi et al. [34] address the general summarization of medical dialogues. They propose combining extractive and abstractive methods that leverage the independent and distinctive local structures formed during a patient's medical history compilation. Liu et al. [35] report a topic-based summarization of general medical domains pertaining to topics such as swelling, headache, chest pain and dizziness. Their encoder-decoder model tries to generate one symptom (topic) at a time. Besides, work on formalizing the conversation text is reported [36]. This work treats the formalizations of the case notes from digital transcripts of the doctor and the patient conversations as a summarization task. Their method involves two steps: prediction of the EHR categories and formal text generation.

Previous works mentioned above either did not focus on aspect-based summarization, or reported on general clinical discussions such as cough, cold and fever. Our work is motivated by the study conducted by Srivastava et al. [37], which reports the summarization-based counseling technique from the therapist and client conversations. They release a conversation dataset which is structured with psychotherapies' core components about symptom and history identification, or the discovery of the patient's behavior. The authors propose an encoder-decoder model based on T5 [38] for their counseling component-guided summarization model. However, a single, generic, summary is being generated in the work and no focus is given on generating aspect-based summaries. Consequently, we extend the work by utilizing the counseling component, viz. symptom and history exploration, patient discovery, and reflections into an aspect-based summarization framework. To this end, we create MentalCLOUDS, a dataset which incorporates summaries aligned with the distinct counseling components. We also explore the efficacy of the state-of-the-art LLMs (encoder-decoder and decoder-only models) for the task of counseling dialogue summarization in this work.

Table 1: Statistics of the MentalCLOUDS dataset (Abbreviations – Dlgs: Dialogues, Utts: Utterances, U/D: Average utterances per dialogue, P Utts: Patient Utterances, T Utts: Therapist Utterances, SH Utts: Symptom and Reasoning Utterances, PD Utts: Patient Discovery Utterances, RT Utts: Reflective Utterances).

| Set | # Dlgs | # Utts | U/D | # P Utts | # T Utts | # SH Utts | # PD Utts | # RT Utts |
|---|---|---|---|---|---|---|---|---|
| Train | 131 | 8342 | 63.68 | 4124 | 4211 | 1882 | 3826 | 884 |
| Val | 21 | 1191 | 56.71 | 594 | 597 | 206 | 445 | 146 |
| Test | 39 | 2010 | 51.53 | 1004 | 1006 | 291 | 1157 | 212 |
| Total | 191 | 11543 | 60.43 | 5722 | 5814 | 2379 | 5428 | 1242 |

Figure 1: Distribution of summary lengths in MentalCLOUDS.

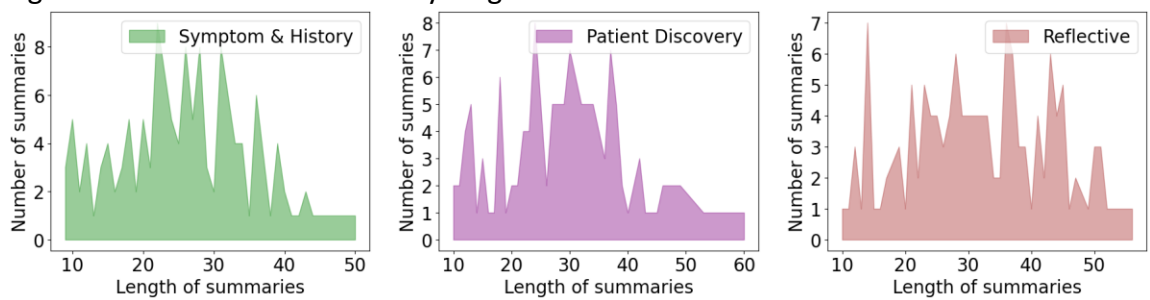

## Methods

### Overview of the Proposed Datasets: MentalCLOUDS

To evaluate the performance of diverse summarization systems across various aspects of counseling interactions, we expand upon the MEMO dataset [37]. Comprising 11.5₹ utterances extracted from 191 counseling sessions involving therapists and patients, this dataset draws from publicly accessible platforms such as YouTube. Embracing a heterogeneous demographic spectrum with distinctive mental health concerns and diverse therapists, the dataset facilitates the formulation of a comprehensive and inclusive approach for researchers. Utilizing pre-processed transcriptions derived from counseling videos, the constituent dialogues within the dataset exhibit a dyadic structure, exclusively featuring patients and therapists as interlocutors. Within each conversation, three pivotal counseling components (aspects) emerge – symptom and history exploration, patient discovery, and reflective utterances.

Our study aims to capture the essence of each aforementioned counseling component, embarking on the creation of three distinct summaries for a single dialogue — each tailored to a specific counseling component. Expanding upon the MEMO dataset, we augment it with annotated dialogue summaries corresponding to the three identified components. Collaborating closely with a team of leading mental health experts (their details are mentioned in the Qualitative Assessment by Experts section), we crafted annotation guidelines and subjected the summary annotations to rigorous validation processes. We call the resultant dataset MentalCLOUDS (Mental health Counseling cOmponent gUided Dialogue Summaries). We highlight its key statistics in Table 1 and Figure 1.

### Data Annotation Process

Conversations in counseling situations can be challenging, given the sensitive nature of the information. A reflective and open attitude of the therapist can facilitate this expression. This dynamic is reinforced by the proposed MentalCLOUDS dataset. This dataset distinguishes the utterances dedicated to symptom exploration, discovering the history of mental health issues and patient behavior, and providing insights into past narratives, thereby shaping the patient's present circumstances. These nuanced elements form the core of our identified counseling components. To improve the richness of the dataset, we collaborated with mental health experts to formulate a set of annotation guidelines [39]. Further, these guidelines serve as a comprehensive framework by which annotators can focus their attention on particular aspects of the conversation that are essential for producing summaries that are customized for each counseling component. By adhering to these guidelines, the therapeutic techniques are captured in the annotations. This ensures the resulting summaries are concise yet rich in informative content for the specific component.

**Psychotherapy Elements:** Within the realm of mental health therapy sessions, distinct counseling components play a pivotal role in facilitating successful interventions. The MentalCLOUDS dataset serves as a valuable resource, furnishing meticulously labeled utterances that encompass three fine-grained components [37]:
- Symptom and History (SH): This facet encapsulates utterances teeming with insightful information crucial for the therapist's nuanced assessment of the patient's situation.
- Patient Discovery (PD): Patients entering counseling sessions often bring intricate thoughts to the fore. Therapists, in turn, endeavor to establish therapeutic connections, creating a conducive environment for patients to articulate and unravel their thoughts. Such utterances by the therapist that encourage patients to reveal their concerns lie in this category.
- Reflecting (RT): Therapists employ concise utterances, allowing ample space for patients to share their life stories and events. Encouraging patient narratives, therapists may also utilize hypothetical scenarios to evaluate actions and enhance understanding.

When crafting a summary for a dialogue *D*, aligned with a specific counseling component *C*, our primary focus rests on utterances marked with *C* within *D* in the MEMO dataset. Consequently, we derive three distinct counseling summaries for each counseling component within a single session to obtain MentalCLOUDS. Table 1 shows the data statistics, where a balanced distribution of patient and therapist utterances within the dataset is evident. Notably, Patient Discovery (PD) emerges as the prevailing label in the dataset, highlighting patient's inclination to discuss ancillary topics rather than focusing solely on their mental health concerns when prompted to share their experiences. In contrast, Reflecting (RT) emerges as the least tagged label in this comprehensive analysis.

**Benchmarking**

In recent years, the spotlight on LLMs has intensified, captivated by their extraordinary performance across diverse applications. From classification tasks like emotion recognition [40] to generative problems such as response generation [41], these models have proven their versatility. In this paper, our focus is directed towards evaluating their capability in the domain of counseling summarisation, specifically using MentalCLOUDS. In our comprehensive analysis, we leverage 11 state-of-the-art pretrained LLM architectures, including a mix of general-purpose and specialized models. These models are considered to carefully assess their performance concerning each facet of counseling component summaries. We explain each of these systems below:

1. **BART** [42]**:** BART (Bidirectional and Auto-Regressive Transformers) is a sequence-to-sequence model designed for various NLP tasks, including text summarization. It employs a transformer architecture with an encoder-decoder structure. It incorporates a denoising autoencoder objective during pretraining, reconstructing the original input from corrupted versions. We use the pretrained base version of the model in our experiments.
2. **T5** [38]**:** T5 (Text-To-Text Transfer Transformer) is a versatile transformer-based model consisting of an encoder-decoder framework with bidirectional transformers. It reframes all NLP tasks as text-to-text tasks, providing a unified approach. T5 learns representations by denoising corrupted input-output pairs. Its encoder captures contextual information while the decoder generates target sequences. The pretrained base version of T5 is employed in our experiments.
3. **GPT-2** [43]**:** GPT-2 (Generative Pre-trained Transformer 2) is a transformer-based language model which comprises a stack of identical layers, each with a multi-head self-attention mechanism and position-wise fully connected feed-forward networks. GPT-2 follows an autoregressive training approach, predicting the next token in a sequence given its context.
4. **GPT-Neo** [44]**:** Trained from the Pile dataset [45], GPT-Neo exhibits a similar architecture as GPT-2 except for a few modifications, such as the usage of local attention in every other layer with a window size of 256 tokens. Additionally, GPT-Neo houses a combination of linear attention [46], a mixture of experts [47], and axial positional embedding [48] to perform comparable to bigger LLMs, like GPT-3.
5. **GPT-J** [49]**:** GPT-J is a transformer model trained using the methodology proposed by Wang et al. [49]. It is a GPT-2-like causal language model trained on the Pile dataset.
6. **FLAN T5** [50]**:** FLAN-T5 is the instruction fine-tuned version of the T5 model with a particular focus on scaling the number of tasks, scaling the model size, and finetuning on chain-of-thought data.

7. **Mistral** [51]**:** Mistral is a decoder-based LM with a sliding window attention mechanism, where it is trained with an 8k context length and fixed cache size, with a theoretical attention span of 128K tokens. Faster inference and lower cache are ensured by using Grouped Query Attention (GQA) [52].
8. **MentalBart** [53]**:** MentalBART is an open-source LLM constructed for interpretable mental health analysis with instruction-following capability. The model is finetuned using the IMHI dataset [53] and is expected to make complex mental health analyses for various mental health conditions.
9. **MentalLlama** [53]**:** Similar to MetalBART, MentalLlama is the counterpart of the LLama architecture but is trained on the mental health IMHI dataset. The model is fine-tuned and is supposed to carry the capability of the LLM with the domain knowledge of mental health.
10. **Llama 2** [54]**:** Llama 2 is an auto-regressive language model that uses an optimized transformer architecture. The tuned versions use Supervised Fine Tuning (SFT) [55] and Reinforcement Learning with Human Feedback (RLHF) [56] to align with human preferences for helpfulness and safety. The model is trained exclusively on publicly available datasets.
11. **Phi 2:** Phi 2 is an extension of Phi 1 [57]. Phi 1 is a transformer-based frugal LLM with the largest variant having 1.3B parameters. It is trained on textbook-quality data. It emphasizes the quality of the data to compensate for its relatively small number of parameters. Phi 2 is a 2.7B parameter model, which shows comparable performances with other larger LLMs despite its smaller size.

## Results

We undertake a comprehensive evaluation of the generated summaries across various architectures, employing a dual approach of quantitative and qualitative assessments.

### Quantitative Assessment

This section reports the aspect-based (psychotherapy element-based) summarization results based on the automatic evaluation scores. Given the generative nature of the task, we employ standard summarization evaluation metrics such as Rouge-1 (R-1), Rouge-2 (R-2), Rouge-L (R-L), and BERTScore (BS) along with their corresponding Precision (P), Recall (R) and F1 scores. Since F1 accounts for Precision and Recall, we compare LLM's performance based on F1 unless stated otherwise. ROUGE (Recall-Oriented Understudy for Gisting Evaluation) [58] assesses the overlap of *n*-grams (sequences of *n* consecutive words) between the generated summary and reference summaries. Specifically, this metric measures the number of overlapping units such as *n*-gram, word sequences, and word pairs between the generated summary evaluated against the gold summary typically created by humans. ROUGE favors the candidate summary with more overlaps with references. This effectively gives more weight to matching *n*-grams occurring in multiple references. This work reports the unigram, bigram ROUGE (viz. ROUGE-1 and ROUGE-2) and ROUGE-L. ROUGE-L takes into account the longest co-occurring n-gram between the candidate and the reference summaries. BERTScore [59] is harnessed to gauge the semantic coherence between the generated summaries and their ground truths. Notably, in the context of counseling summaries, which are inherently tied to a domain-specific conversation, we embark on a meticulous qualitative examination of the generated summaries for individual counseling components.

**Symptom and History (SH):** Table 2 reports the automatic evaluation scores of LLMs on the summarization task for the Symptom and History (SH) psychotherapy element. MentalLlama outperforms the other LLMs across all the automatic evaluation metrics. For the R-1 metric, it achieves an F1 score of 30.86, followed by MentalBART with an F1 score of 28.00. In terms of the R-2 metric, Mistral is comparable with MentalLlama with a difference of mere 0.90 F1 score. Similarly, for R-L, Mistral is preceded by MentalLlama by a difference of 2.93 F1 score.

Table 2. Results obtained on MentalCLOUDS for the summarization task on the Symptom and History (SH) psychotherapy element. Best results are marked in bold.

| | R-1 | | | R-2 | | | R-L | | | BS | | |
|---|---|---|---|---|---|---|---|---|---|---|---|---|
| **Model** | P | R | F | P | R | F | P | R | F | P | R | F |
| BART | 12.91 | 28.84 | 16.26 | 1.88 | 5.07 | 2.47 | 10.21 | 23.97 | 13.19 | 85.81 | 85.81 | 85.81 |
| T5 | 22.16 | 19.81 | 19.74 | 2.18 | 1.78 | 1.85 | 16.12 | 14.51 | 14.36 | 85.38 | 85.38 | 85.38 |
| MentalBART | 30.31 | 29.02 | 28.00 | 6.06 | 5.29 | 5.46 | 20.85 | 20.34 | 19.4 | 88.34 | 88.34 | 88.34 |
| Flan-T5 | 21.45 | **33.15** | 24.80 | 3.84 | 6.08 | 4.54 | 17.15 | 26.53 | 19.76 | 86.94 | 86.94 | 86.94 |
| GPT2 | 6.59 | 14.62 | 8.91 | 1.06 | 2.34 | 1.42 | 5.12 | 11.37 | 6.93 | 83.65 | 83.65 | 83.65 |
| GPT-Neo | 9.97 | 19.91 | 13.01 | 1.01 | 2.30 | 1.38 | 7.89 | 15.91 | 10.33 | 83.12 | 83.12 | 83.12 |
| GPT-J | 13.22 | 29.99 | 17.88 | 3.37 | **7.96** | 4.59 | 10.71 | 24.34 | 14.47 | 86.28 | 86.28 | 86.28 |
| MentalLlama | **33.03** | 32.79 | **30.86** | **8.66** | 6.5 | **7.28** | **27.73** | **27.3** | **29.55** | **89.4** | **90.99** | **90.99** |
| Mistral | 29.07 | 26.56 | 25.41 | 7.03 | 5.2 | 7.19 | 25.45 | 25.61 | 26.62 | 83.42 | 85.96 | 83.05 |
| Llama-2 | 28.49 | 24.17 | 23.47 | 6.4 | 4.68 | 6.63 | 22.7 | 23.04 | 23.66 | 82.86 | 83.8 | 81.62 |
| Phi-2 | 21.23 | 10.42 | 13.81 | 1.89 | 1.43 | 1.78 | 14.56 | 9.19 | 11.26 | 84.25 | 82 | 83.11 |

**Patient Discovery (PD):** The experimental results presented in Table 3 focus on the summarization task for the Patient Discovery (PD) psychotherapy element. Considering the R-1 metric, MentalLlama demonstrates superior performance compared to other LLMs. MentalLlama shows a 30.95 F1 score, followed by MentalBART (29.94 F1 Score). For the R-2 metric, GPT-J outperforms the other models, followed by MentalLlama. Additionally, in terms of the R-L metric, the top two highest F1 score models are MentalLlama and Mistral. Finally, MentalBART supersedes the other models with an F1 score of 88.61 w.r.t BS metric. Overall, the score indicates that LLMs such as MentalLlama and MentalBART, which were pre-trained on the mental domain data, show consistent superiority. Notably, the base Mistral model also performs comparable and sometimes better than the models trained on the mental health domain data.

Table 3. Results obtained on MentalCLOUDS for the summarization task on the Patient Discovery (PD) psychotherapy element. Best results are marked in bold.

| | R-1 | | | R-2 | | | R-L | | | BS | | |
|---|---|---|---|---|---|---|---|---|---|---|---|---|
| **Model** | P | R | F | P | R | F | P | R | F | P | R | F |
| BART | 20.82 | 43.24 | 26.72 | 5.97 | 12.93 | 7.74 | 16.38 | 34.82 | 21.14 | 87.35 | 87.35 | 87.35 |
| T5 | 9.43 | 47.29 | 15.34 | 3.03 | 16.9 | 5.01 | 8.39 | 42.58 | 13.67 | 84.77 | 84.77 | 84.77 |
| MentalBART | 33.51 | 29.94 | 29.94 | 9.36 | 7.94 | 8.06 | 23.39 | 21.44 | 21.1 | **88.61** | 88.61 | **88.61** |
| Flan-T5 | 21.08 | 35.61 | 24.44 | 4.81 | 8.89 | 5.63 | 16.13 | 28.29 | 18.94 | 86.52 | 86.52 | 86.52 |
| GPT2 | 13.66 | 36.24 | 19.57 | 4.08 | 11.27 | 5.94 | 10.93 | 29.42 | 15.7 | 85.21 | 85.21 | 85.21 |
| GPT-Neo | 12.96 | 29.93 | 17.83 | 2.32 | 5.44 | 3.22 | 9.84 | 23.1 | 13.6 | 82.72 | 82.72 | 82.72 |
| GPT-J | 19.78 | **53.33** | 28.85 | **12.68** | **35.71** | **18.71** | 16.12 | **43.33** | 23.49 | 86.43 | 86.43 | 86.43 |
| MentalLlama | **24.56** | 43.84 | **30.95** | 9.55 | 26.01 | 12.79 | **23.77** | 38.98 | **29.17** | 84.63 | **88.95** | 86.68 |
| Mistral | 22.84 | 39.02 | 27.54 | 8.78 | 25.79 | 11.35 | 21.9 | 35.98 | 24.02 | 86.62 | 87.28 | 84.49 |
| Llama-2 | 20.22 | 34.7 | 26.1 | 8.41 | 21.13 | 10.39 | 14.73 | 21.44 | 17.79 | 78.81 | 88.06 | 81.48 |
| Phi-2 | 18.72 | 9.23 | 12.45 | 5.61 | 4.44 | 4.96 | 13.94 | 8.73 | 10.98 | 84.25 | 82 | 80.05 |

Table 4. Results obtained on MentalCLOUDS for the summarization task on the Reflecting (RT) psychotherapy element. Best results are marked in bold.

| | R-1 | | | R-2 | | | R-L | | | BS | | |
|---|---|---|---|---|---|---|---|---|---|---|---|---|
| **Model** | P | R | F | P | R | F | P | R | F | P | R | F |
| BART | 17.01 | 23.04 | 18.08 | 2.87 | 4.25 | 3.22 | 12.68 | 17.79 | 13.66 | 85.26 | 85.26 | 85.26 |
| T5 | 34.13 | 19.32 | 24.31 | 7.21 | 3.97 | 5.04 | 22.95 | 12.82 | 16.21 | 84.92 | 84.92 | 84.92 |
| MentalBART | **34.99** | 36.54 | 34.46 | **10.24** | 10.66 | 10.07 | 24.52 | 25.8 | 24.25 | **88.7** | **88.7** | **88.7** |
| Flan-T5 | 25.1 | 41.4 | 30.15 | 7.19 | 12.03 | 8.64 | 18.52 | 31 | 22.36 | 87.41 | 87.41 | 87.41 |
| GPT2 | 2.84 | 7.54 | 4.08 | 0.14 | 0.33 | 0.2 | 2.35 | 6.34 | 3.39 | 82.66 | 82.66 | 82.66 |
| GPT-Neo | 1.14 | 3.97 | 1.74 | 0 | 0 | 0 | 1.14 | 3.97 | 1.74 | 80.88 | 80.88 | 80.88 |
| GPT-J | 17.6 | 38.33 | 23.71 | 5.07 | **13.04** | 7.13 | 14.98 | 32.85 | 20.18 | 86.94 | 86.94 | 86.94 |
| MentalLlama | 31.68 | **54.76** | **39.52** | 8.26 | 11.99 | **10.17** | **27.13** | **37.59** | **26.56** | 84.77 | 86.92 | 87.43 |

| | | | | | | | | | | | |
|---|---|---|---|---|---|---|---|---|---|---|---|
| Mistral | 29.15 | 49.28 | 38.33 | 8.42 | 11.87 | 8.34 | 24.41 | 34.2 | 23.44 | 78.83 | 79.97 | 84.81 |
| Llama-2 | 26.93 | 43.81 | 31.22 | 6.1 | 9.23 | 8.24 | 16.82 | 20.67 | 16.21 | 78.93 | 86.05 | 82.19 |
| Phi-2 | 10.61 | 5.21 | 6.91 | 0.94 | 0.71 | 0.89 | 7.28 | 4.6 | 5.53 | 86.94 | 82.17 | 84.49 |

**Reflecting (RT):** Table 4 reports the automatic evaluation scores on the summarization task for the Reflecting (RT) psychotherapy element. Considering the R-1 metric, MentalLlama and Mistral are the best two models, with 39.52 and 38.33 F1 scores, respectively. Similarly, MentalLlama demonstrates its superiority against other LLMs in terms of the R-2, R-L and BS metrics. Moreover, the scores of summarization tasks for this psychotherapy element are analogous to that of the previous two summarization tasks viz. SH and PD, wherein the mental-health-specific LLMs exhibit their superiority over the other LLMs.

### Qualitative Assessment by Experts

In order to conduct a comprehensive expert assessment, five healthcare professionals were employed to assess the clinical appropriateness of the summaries produced by the LLMs based on the evaluation framework of Sekhon et al. [39]. Among them were two clinical psychologists, with the remaining three comprising psychiatrists and medical practitioners. Of the group, four were male, and one was female, with ages ranging from 40 to 55 years and possessing over a decade of therapeutic experience.

The evaluation framework encompasses six crucial parameters — affective attitude, burden, ethicality, coherence, opportunity costs, and perceived effectiveness. Experts evaluate each summary against these acceptability parameters, assigning continuous ratings on a scale from 0 to 2, where a higher rating signifies enhanced acceptability. Additionally, we incorporate a new parameter – the extent of hallucination. It is categorical – 0 (*too much hallucinated*), 1 (*hallucination barely observed*), and 2 (*no hallucination observed*). These evaluative dimensions are defined in Table 5.

Table 5. Explanation of the experts' evaluation metrics as presented in [36].

| Construct | Definition | Application |
|---|---|---|
| Affective attitude | How an individual feels about an intervention. | What are your perceptions of the summarisation based upon your clinical knowledge |
| Burden | Perceived amount of effort required to participate | How much effort is required to understand the summarisation, consider: spelling, grammar, overall interpretation |
| Ethicality | Extent to which this is a good fit with your organization's value system | How does this align with your respective code of ethics? Are there concerns? |
| Intervention Coherence | Extent to which the intervention is understood | How well the summaries are understood |
| Opportunity Cost | The extent to which one would benefit from using this intervention | Pros and cons of using this intervention in your respective setting |
| Perceived Effectiveness | Extent to which this intervention will perform in the intended setting | How well this will perform in your clinical setting |
| Hallucination extent | Extent to which this intervention is hallucinated | The generated text is incorrect, nonsensical or contains global information apart from the context of the conversation. |

Table 6. Qualitative evaluation by human experts. These scores are averaged out from the five expert raters. (Abbreviations – AA: Affective Attitude, BD: Burden, ET: Ethicality, IC: Intervention Coherence, OC: Opportunity Cost, PE: Perceived Effectiveness). The variances among the raters' scores are also shown. Best results are marked in bold.

| Model | | AA | BD | ET | IC | OC | PE |
|---|---|---|---|---|---|---|---|
| Mistral | Average | **1.12** | **1.33** | **1.42** | **1.13** | **0.98** | **0.90** |
| | Variance | 0.22 | 0.10 | 0.14 | 0.20 | 0.22 | 0.26 |
| MentalLlama | Average | **1.12** | **1.33** | 1.36 | 1.06 | 0.94 | 0.88 |
| | Variance | 0.14 | 0.05 | 0.10 | 0.13 | 0.15 | 0.20 |
| MentalBART | Average | 0.95 | 1.28 | 1.33 | 1.01 | 0.84 | 0.76 |
| | Variance | 0.08 | 0.02 | 0.13 | 0.05 | 0.11 | 0.16 |

Table 6 reports the clinical experts' scores averaged over the ratings given by five experts. The clinical acceptability framework [39] involves six parameters – affective attitude, burden, ethicality, coherence, opportunity costs, and perceived effectiveness (c.f. Table 5 for more details). We select the three best LLMs (MentalLlama, Mistral and MentalBART) for the expert evaluation based on the automatic result. Notably, Mistral outperforms the other two LLMs in all the metrics even though the other two LLMs were finetuned in the mental health domain. Overall, all the raters were more aligned in rating the MentalBART model with lesser variance as compared to the other two LLMs for all the metrics. However, all three LLMs give higher ratings on the surface-level characteristic metric (Burden) or subjective metric (Affective Attitude) as compared to the Opportunity Cost and efficacy (Perceived Effectiveness). As all three models have poor scores on the more sensitive aspects i.e. the overall efficacy and the opportunity cost, this indicates that these models share the same weakness and are not suitable for clinical use as they stand now.

Table 7. Hallucination frequency marked by experts for the top three LLMs. Here the average of hallucination frequencies for each rater are reported.

| Hallucination Type | Mistral | MentalLlama | MentalBART |
|---|---|---|---|
| No Hallucination Observed | 29.3 | 30.3 | 29.7 |
| Hallucination Barely Observed | 5.1 | 5.6 | 7.3 |
| Too much Hallucinated | 4.3 | 3 | 2 |

**Extent of Hallucination:** Additionally, the evaluation of hallucination identification is divided into three categories: *no hallucination* observed, *hallucination barely observed*, and *too much hallucinated* in a set of 39 conversations. These categories essentially determine how well the response is consistent with the context and whether it is also incorrect, nonsensical, or contains global information beyond the scope of the conversation. The results are summarized in [Table 7](). Out of the 39 test conversations, the majority of cases on average demonstrate *no hallucination*, with Mistral and MentalBART achieving 75.13% and 76.15% each, while MentalLlama shows a slightly higher value of 77.69%. Among those samples where *hallucination barely observed* is reported, all three models fall into almost the same range – Mistral and MentalLlama account for 13.08% and 14.36% each, and MentalBert shows a somewhat elevated value of 18.72%. Notably, the models turn out to be lower in the *too much hallucinated* group – Mistral with only 11.03%, MentalLlama with only 7.69%, and MentalBART with about 5.13%. This information confirms the capability of these AI models to follow faithfully whenever there is no hallucination and further highlights more subtle degrees of hallucination detection in different tasks they were tested on.

The results are consistently adequate in all three models, and there is more or less an equal division of the types of hallucinations observed by different raters. Importantly, all three models show a large number of cases with no hallucinations that could be deemed reliable performance, implying their ability to retain fidelity to the original content. [Table 7]() presents the total numbers for each model and hallucination category as assessed by five individual raters. The data shows fluctuations in how hallucinations are perceived among different models and stresses the importance of reviewing evaluations from numerous appraisers for a complete assessment.

# Discussion

## Principal Findings

In this work, we assessed the state-of-the-art LLMs on the aspect-based summarization task of mental health therapy conversations. These therapy conversations are long and require a good amount of effort to gain insights. In our premise, we summarized these long conversations, thereby reducing the efforts of experts. We further proposed MentalCLOUDS with aspect-based summaries for each conversation.

Specifically, we benchmarked 11 LLMs for aspect-based summarization and evaluated them using both automatic and human evaluation approaches. The automatic score revealed the superiority of the LLMs trained on mental domain data. Two domain-specific LLMs, MentalLlama and MentalBART, consistently outperformed the rest of the LLMs across all aspects. Notably, although Mistral is not specifically trained in the mental health domain, its scores are comparable with MentalLlama, the overall best-performing model.

This work also showcased the prowess of decoder-only LLMs instead of strong encoder-decoder-based LLMs. Typically, encoder-decoder models favor sequence-to-sequence tasks such as summarization, where a sequence of input texts is mapped to a sequence of output text. On the contrary, the decoder-based model, i.e., MentalLlama and Mistral, consistently outperformed the other encoder-decoder models such as BART, T5 and Flan-T5. The only exception is MentalBART since it is finetuned on the mental health dataset.

The counseling dataset was curated from multiple multimedia online sources such as youtube transcripts [37]. Hence, most of these natural conversations are incoherent and grammatically unfluent. Even with these imperfections, the LLMs were mostly able to construct meaningful summaries that contained coherent narratives with a clear beginning and end. However, the models did not do as well with the structure separation of the information. The sections of "symptoms and history", "patient discovery", and "reflection" frequently overlap, posing clinical and legal problems. History is considered clinically sacrosanct and should not be contaminated by the therapist's interpretation. History is also citable in legal cases as client evidence, while interpretations are not. The models are also unable to identify psychotherapy types (e.g., cognitive behavior therapy) and therapy techniques, which form an integral part of counseling notes. For example, when subjects are engaged using a motivational interviewing framework, the essential processes and their outcomes that a human summarizer would record failed to find a place in the summaries. Important negative histories gathered during the session, such as the history of suicide risk or substance use were also not recorded, and in at least one instance, the presence of suicide risk was not identified. In general, the models exhibited stronger performance in handling medical histories and examinations but struggled when faced with more technical and sensitive aspects, such as conversations related to actual therapeutic strategies.

## Limitations

It is crucial to address the limitations of this study for a comprehensive understanding. First, this work aims to benchmark the efficacy of only 11 LLMs on the aspect-based summarization task. Second, for faster and easier reproduction, we did not assess models larger than 7 billion parameters; however, such models can be part of future examinations. Third, for the initial study and to promote research in this field, only open-source models were assessed in this work. However, inspecting closed models such as ChatGPT, Claude and Gemini can be an interesting future research avenue. Finally, this work explored only three aspects (counseling component) of the conversation. However, conversations are subjective and can have more than three components. Additionally, the counseling sessions in this work represented a certain demographic region (American) and thus may not apply to therapy counseling for other demographics.

## Conclusions

Our study benchmarked the efficacy and role of large language models towards component-guided counseling summarization tasks. In doing so, we introduced a new dataset, MentalCLOUDS, which comprises summaries corresponding to three counseling components. Experimental results confirmed the superiority of the LLMs finetuned in the mental domain (MentalLlama and MentalBART) against the out-of-the-box LLMs. Notably, the out-of-the-box Mistral model seems comparable to and sometimes better than the mental-domain finetuned LLMs. However, as per the experts' evaluation, these LLMs often failed to distinguish between the counseling components during the summary generation. Overall, these models excelled in managing medical histories and examinations but faced challenges with technical and sensitive aspects, such as therapy conversations, thereby limiting their clinical utility as they stand now.

## Data Availability

The dataset will be available upon request.

## Conflicts of Interest

The authors do not have any conflict of interest.

## Abbreviations

AI: Artificial Intelligence
BART: Bidirectional and Auto-Regressive Transformers
BERT: Bidirectional Encoder Representations from Transformers
CBT: Cognitive Behavioural Therapy
DL: Deep Learning
GPT: Generative Pre-trained Transformer
JMIR: Journal of Medical Internet Research
LLM: Large Language Model
LSTM: Long Short Term Memory
MentalCLOUDS: Mental Health Counseling Component Guided Dialogue Summaries
PD: Patient Discovery

PHQ: Patient Health Questionnaire
PICO: Patient/Problem, Intervention, Comparison, and Outcome
RNN: Recurrent Neural Network
ROUGE: Recall-Oriented Understudy for Gisting Evaluation
BiRNN: Bidirectional Neural Network
RT: Reflecting
SH: Symptom and History
T5: Text-To-Text Transfer Transformer